\let\accentvec\vec
\documentclass[runningheads]{llncs}

\let\vec\accentvec
\usepackage[T1]{fontenc}
\usepackage{graphicx}
\usepackage{comment}
\usepackage{amsmath,amssymb} 
\usepackage{color}

\usepackage{booktabs}
\usepackage{multirow}
\usepackage{array}
\usepackage{subfigure}
\usepackage{siunitx}
\newcolumntype{L}[1]{>{\raggedright\let\newline\\\arraybackslash}m{#1}}
\newcolumntype{C}[1]{>{\centering\let\newline\\\arraybackslash}m{#1}}

\begin{document}
\pagestyle{headings}
\mainmatter
\def\ECCVSubNumber{2538}  

\title{Yet Another Intermediate-Level Attack} 

\titlerunning{Yet Another Intermediate-Level Attack}
%
\author{Qizhang Li\inst{1}\thanks{Work done during an internship at ByteDance AI Lab, under the guidance of Yiwen Guo$^\dagger$ who is the corresponding author.} \and  
Yiwen Guo\inst{1}$^\dagger$ \and
Hao Chen\inst{2}}
\authorrunning{Q. Li$^\star$, Y. Guo$^\dagger$, H. Chen}
%
\institute{ByteDance AI Lab \and University of California, Davis \\ 
\email{\{liqizhang,guoyiwen.ai\}@bytedance.com}\quad \email{chen@ucdavis.edu}\\
}
\maketitle

\begin{abstract}
The transferability of adversarial examples across deep neural network (DNN) models is the crux of a spectrum of black-box attacks. In this paper, we propose a novel method to enhance the black-box transferability of baseline adversarial examples. By establishing a linear mapping of the intermediate-level discrepancies (between a set of adversarial inputs and their benign counterparts) for predicting the evoked adversarial loss, we aim to take full advantage of the optimization procedure of multi-step baseline attacks. We conducted extensive experiments to verify the effectiveness of our method on CIFAR-100 and ImageNet. Experimental results demonstrate that it outperforms previous state-of-the-arts considerably. Our code is at https://github.com/qizhangli/ila-plus-plus.
\end{abstract}

\section{Introduction}

The adversarial vulnerability of deep neural networks (DNNs) has been extensively studied over the years~\cite{Szegedy2014,Goodfellow2015,Moosavi2016,CW2017,Madry2018,Athalye2018obfuscated,Guo2018sparse,Guo2020on}. 
It has been demonstrated that intentionally crafted perturbations, that are small enough to be imperceptible to human eyes, on a natural image can fool advanced DNNs to make arbitrary (incorrect) predictions.
Along with this intriguing phenomenon, it is also pivotal that the adversarial examples crafted on one DNN model can fail another with a non-trivial success rate~\cite{Szegedy2014,Goodfellow2015}.
Such a property, called the transferability (or generalization ability) of adversarial examples, plays a vital role in many black-box adversarial scenarios~\cite{Papernot2016transferability,Papernot2017}, where the architecture and parameters of the victim model is hardly accessible.

Endeavors have been devoted to studying the transferability of adversarial examples. 
Very recently, intermediate-layer attacks~\cite{Zhou2018,Inkawhich2019,Huang2019} have been proposed to improve the transferability. It was empirically shown that larger mid-layer disturbance (in feature maps) leads to higher transferability in general. 
In this paper, we propose a new method for improving the transferability of adversarial examples generated by any baseline attack, just like in~\cite{Huang2019}. 
Our method operates on the mid-layer feature maps of a source model as well.
It attempts to take full advantage of the directional guides gathered at each step of the baseline attack, by maximizing the scalar projection on a spectrum of intermediate-level discrepancies.
The effectiveness of the method was testified on a variety of image classification models on CIFAR-100~\cite{Krizhevsky2009} and ImageNet~\cite{Russakovsky2015}, and we show that it outperforms previous state-of-the-arts considerably.

\section{Related Work}

Adversarial attacks can be categorized into white-box attacks and black-box attacks, according to how much information of a victim model is leaked to the adversary~\cite{Papernot2016transferability}. Initial attempts of performing black-box attacks rely on the transferability of adversarial examples~\cite{Papernot2016transferability,Papernot2017,Liu2017}. Despite the excitement about the possibility of performing attacks under challenging circumstances, early transfer-based methods often suffer from low success rates, and thus an alternative trail of research that estimates gradient from queries also becomes prosperity~\cite{Chen2017,Ilyas2018,Ilyas2019,Guo2019,Bhagoji2018,Tu2019,Brendel2018,Yan2019}.
Nevertheless, there exist applications where queries are difficult and costly to be issued to the victim models, and it is also observed that some stateful patterns can be detected in such methods~\cite{Chen2019stateful}. 

Recently, a few methods have been proposed to enhance the transferability of adversarial examples, boosting the transfer-based attacks substantially.
They show that maximizing disturbance in intermediate-level feature maps instead of the final cross-entropy loss delivers higher adversarial transferability.
To be more specific, Zhou et al.~\cite{Zhou2018} proposed to maximize the discrepancy between an adversarial example and its benign counterpart on DNN intermediate layers and simultaneously reduce spatial variations of the obtained results.
Requiring a target example in addition, Inkawhich et al.~\cite{Inkawhich2019} also advocated performing attacks on the intermediate layers. 
The most related work to ours comes from Huang et al.~\cite{Huang2019}.
Their method works by maximizing the scalar projection of the adversarial example onto a guided direction (which can be obtained by performing one of many off-the-shelf attacks~\cite{Goodfellow2015,Kurakin2017,Madry2018,Dong2018,Zhou2018}) beforehand, on a specific intermediate layer. 
Our method is partially motivated by Huang et al.'s~\cite{Huang2019}. 
It is also proposed to enhances the adversarial transferability, yet our method takes the whole optimization procedure of the baseline attacks rather than their final results as guidelines. As will be discussed, we believe temporary results probably provide more informative and more transferable guidance than the final result of the baseline attack. 
The problem setting will be explained in the following subsection.

\subsection{Problem Setting}\label{sec:problem}
\begin{figure}[ht]
	\centering \vskip -0.2in
	\includegraphics[width=0.99\textwidth]{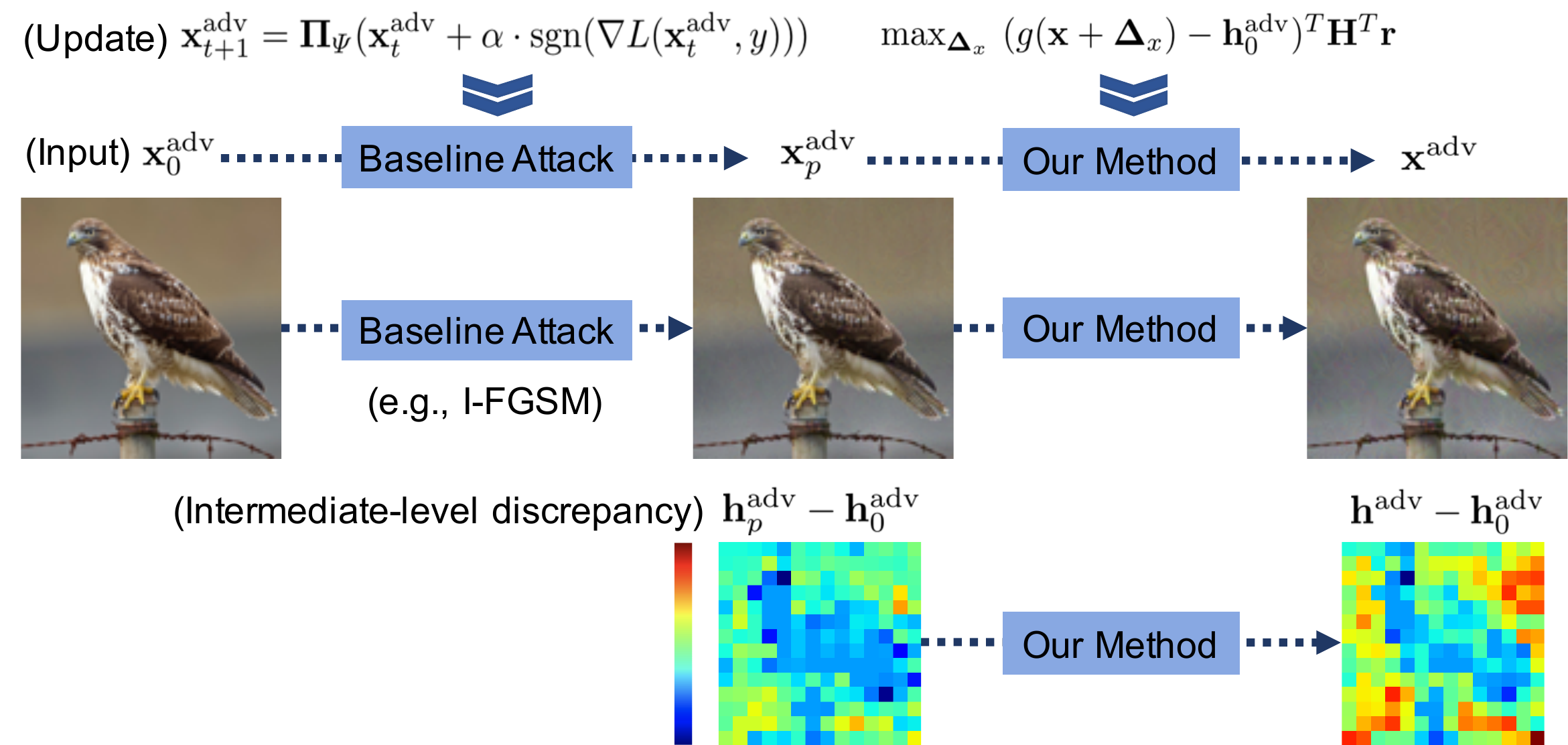} \vskip -0.15in
	\caption{Pipeline of our method for enhancing the black-box transferability of adversarial examples, which is comprised of two sequential phases, one for performing the baseline attack (e.g., I-FGSM~\cite{Kurakin2017}, PGD~\cite{Madry2018}, MI-FGSM~\cite{Dong2018}, etc) and the other for enhancing the baseline result $x^{\mathrm{adv}}_p$. In particular, the chartreuse-yellow background in $\mathbf h_p^{\mathrm {adv}}-\mathbf h_0^{\mathrm {adv}}$ on the left heatmap indicates a much lower disturbance than that in $\mathbf h^{\mathrm {adv}}-\mathbf h_0^{\mathrm {adv}}$ on the right. The discrepancies of feature maps are illustrated from a spatial size of $14\times14$.} \vskip -0.28in
\label{fig:pipeline}
\end{figure}

In this paper, we focus on enhancing the transferability of off-the-shelf attacks, just like Huang et al.'s intermediate-level attack (ILA)~\cite{Huang2019}. We mostly consider multi-step attacks which are generally more powerful on the source models. Suppose that a basic iterative FGSM (I-FGSM) is performed a priori as the \textbf{baseline attack}, we have 
\begin{equation}\label{eq:ifgsm}
    \mathbf x^{\mathrm{adv}}_{t+1} = \mathbf \Pi_\Psi(\mathbf x^{\mathrm{adv}}_t + \alpha\cdot \mathrm{sgn}(\nabla L(\mathbf x^{\mathrm{adv}}_t, y))),
\end{equation}
in which $\Psi$ is a presumed valid set for the adversarial examples and $\mathbf \Pi_\Psi$ denotes a projection onto the set, given $\mathbf x^{\mathrm{adv}}_0=\mathbf x$ and its original prediction $y$. The typical I-FGSM performs attacks after running Eq.~\eqref{eq:ifgsm} for $p$ times to obtain the final adversarial example $\mathbf x^{\mathrm{adv}}_p$. We aim to improve the success rate of the generated adversarial example on some victim models whose architecture and parameters are unknown to the adversary. As depicted in Fig.~\ref{fig:pipeline}, the whole pipeline consists of two phases. The first phase is to perform the baseline attack just as normal, precursor to \textbf{the enhancement phase} where our method or ILA can be applied.

\section{Our Method}\label{sec:method}

As has been mentioned, adversarial attacks are mounted by maximizing some prediction loss, e.g., the cross-entropy loss~\cite{Goodfellow2015,Madry2018} and the hinged logit-difference loss~\cite{CW2017}. The applied prediction loss, which is dubbed \textbf{adversarial loss} in this paper, describes how likely the input shall be mis-classified by the current model.
For introducing our method, we will first propose a new objective function that utilizes the temporary results $\mathbf x^{\mathrm{adv}}_0\ldots \mathbf x^{\mathrm{adv}}_t \ldots \mathbf x^{\mathrm{adv}}_{p-1}$ as well as the final result $\mathbf x^{\mathrm{adv}}_{p}$ of a multi-step attack that takes $p+1$ optimization steps in total, e.g., I-FGSM whose update rule is introduced in Eq.~\eqref{eq:ifgsm}. 

We also advocate mounting attacks on an intermediate layer of the source model, just like prior arts~\cite{Zhou2018,Inkawhich2019,Huang2019}.
Concretely, given $\mathbf x^{\mathrm{adv}}_t$ as a (possibly adversarial) input, we can get the mid-layer output $\mathbf h^{\mathrm{adv}}_t=g(\mathbf x^{\mathrm{adv}}_t)\in\mathbb R^m$ and the adversarial loss $l_t:=L(\mathbf x^{\mathrm{adv}}_t, y)$ from the source model with $L(\cdot, \cdot)$, at a specific intermediate layer.
With a multi-step baseline attack running for a sufficiently long period of time, we can collect a set of \textbf{intermediate-level discrepancies} (i.e., ``perturbations'' of feature maps) and adversarial loss values $\{(\mathbf h_t^{\mathrm{adv}}-\mathbf h^{\mathrm{adv}}_0, l_t)\}$, and further establish a direct mapping of the intermediate-level discrepancies to predicting the adversarial loss.
For instance, a linear (regression) model can be obtained by simply solving a regularized problem. 
\begin{equation}\label{eq:lr}
    \min_{\mathbf w}\ \sum^p_{t=0} (\mathbf w^T (\mathbf h^{\mathrm{adv}}_t-\mathbf h^{\mathrm{adv}}_0) - l_t)^2 + \lambda \|\mathbf w\|^2,
\end{equation}
in which $\mathbf w\in \mathbb R^m$ is the parameter vector to be learned. The above optimization problems can be written in a matrix/vector form: $\min_{\mathbf w} \|\mathbf r - \mathbf H\mathbf w\|^2+\lambda \|\mathbf w\|^2$, in which the $t$-th row of $\mathbf H\in\mathbb R^{(p+1)\times m}$ is $(\mathbf h_t^{\mathrm{adv}}-\mathbf h^{\mathrm{adv}}_0)^T$ and the $t$-th entry of $\mathbf r\in\mathbb R^{p+1}$ is $l_t$. The problem has a closed-form solution:
$\mathbf w^\ast = (\mathbf H^T \mathbf H + \lambda \mathbf I_m)^{-1} \mathbf H^T \mathbf r$.

Rather than maximizing the conventional cross-entropy loss as in FGSM~\cite{Goodfellow2015}, I-FGSM~\cite{Kurakin2017}, and PGD~\cite{Madry2018}, we opt to optimizing 
\begin{equation}\label{eq:opt0}
    \max_{\mathbf{\Delta}_x}\ (g(\mathbf x+\mathbf{\Delta}_x) - \mathbf h^{\mathrm{adv}}_0)^T \mathbf w^\ast, \quad \mathrm{s.t.}\ (\mathbf x+\mathbf{\Delta}_x) \in \Psi
\end{equation}
to generate pixel-level perturbations with maximum \emph{expected} adversarial loss in the sense of the established mapping from the feature space to the loss space.
Both one-step (e.g., FGSM) and multi-step algorithms (e.g., I-FGSM and PGD) can be used to naturally solve the optimization problem~\eqref{eq:opt0}. Here we mostly consider the multi-step algorithms, and as will be explained, our method actually boils down to ILA~\cite{Huang2019} in a one-step case. Note that the intermediate-level feature maps are extremely high dimensional. The matrix $(\mathbf H^T \mathbf H + \lambda \mathbf I_m)\in \mathbb R^{m\times m}$ thus becomes very high dimensional as well, and calculating its inverse is computational demanding, if not infeasible. 
While on the other hand, multi-step baseline attacks only update for tens or at most hundreds of iterations in general, and we have $p\ll m$.
Therefore, we utilize the Woodbury identity
\begin{equation}\label{eq:woodbury}
\begin{aligned}
    \mathbf H^T \mathbf H + \lambda \mathbf I_m &= \frac{1}{\lambda}I - \frac{1}{\lambda^2}\mathbf H^T(\frac{1}{\lambda}\mathbf H \mathbf H^T + \mathbf I_p)^{-1} \mathbf H \\
    &=\frac{1}{\lambda}I - \frac{1}{\lambda}\mathbf H^T(\mathbf H \mathbf H^T + \lambda\mathbf I_p)^{-1} \mathbf H
\end{aligned}
\end{equation}
so as to calculate the matrix inverse of $(\mathbf H \mathbf H^T + \lambda\mathbf I_p)$ instead, for gaining higher computational efficiency. We can then rewrite the derived optimization problem in Eq.~\eqref{eq:opt0} as 
\begin{equation}\label{eq:opt1}
\begin{aligned}
    \max_{\mathbf{\Delta}_x}&\ (g(\mathbf x+\mathbf{\Delta}_x) - \mathbf h^{\mathrm{adv}}_0)^T (\mathbf I_p-\mathbf H^T(\mathbf H \mathbf H^T + \lambda\mathbf I_p)^{-1} \mathbf H) \mathbf H^T \mathbf r, \\ \mathrm{s.t.}&\ (\mathbf x+\mathbf{\Delta}_x) \in \Psi.
\end{aligned}
\end{equation}

It is worth mentioning that, with a drastically large ``regularizing" parameter $\lambda$, we have $\mathbf H^T(\mathbf H \mathbf H^T + \lambda\mathbf I)^{-1} \mathbf H\approx \mathbf 0$ and, in such a case, the optimization problem in Eq.~\eqref{eq:opt1} approximately boils down to: 
$\max_{\mathbf{\Delta}_x}\ (g(\mathbf x+\mathbf{\Delta}_x) - \mathbf h^{\mathrm{adv}}_0)^T \mathbf H^T \mathbf r$. If only the intermediate-level discrepancy evoked by the final result $\mathbf x^{\mathrm{adv}}_p$ along with its corresponding adversarial loss is used in the optimization (or a single-step baseline attack is applied), the optimization problem is mathematically equivalent to that considered by Huang et al.~\cite{Huang2019}, making their ILA a special case of our method. 
In fact, the formulation of our method suggests a maximized projection on a linear combination of the intermediate-level discrepancies, which are derived from the temporary results $\mathbf x^{\mathrm{adv}}_0\ldots \mathbf x^{\mathrm{adv}}_t \ldots \mathbf x^{\mathrm{adv}}_{p-1}$ and the final result $\mathbf x^{\mathrm{adv}}_p$ of the multi-step baseline attack.
Since the temporary results 
possibly provide complementary guidance to the final result, our method can be more effective. 

In~\eqref{eq:opt0} and~\eqref{eq:opt1}, we encourage the perturbation $g(\mathbf x+\mathbf{\Delta}_x) - \mathbf h^{\mathrm{adv}}_0$ on feature maps to align with $\mathbf w^\ast$, to gain more powerful attacks on the source model. In the meanwhile, the magnitude of the intermediate-level discrepancy $\|g(\mathbf x+\mathbf{\Delta}_x) - \mathbf h^{\mathrm{adv}}_0\|$ is anticipated to be large to improve the transferability of the generated adversarial examples, as also advocated in ILA. Suppose that we are given two directional guides that would lead to similar adversarial loss values on the source model, yet remarkably different intermediate-level disturbance via optimization using for instance ILA. 
One may anticipate the one that causes larger disturbance in an intermediate layer to show better black-box transferability. Nevertheless, it is not guaranteed that the final result of the baseline attack offers an exciting prospect of achieving satisfactory intermediate-level disturbance in the followup phase. 
By contrast, our method endows the enhancement phase some capacities of exploring a variety of promising directions and their linear combinations that trade off the adversarial loss on the source model and the black-box transferability.
Experimental results in Section~\ref{sec:sota} shows that our method indeed achieves more significant intermediate-level disturbance in practice.

\subsection{Intermediate-level Normalization}
In practice, the intermediate-level discrepancies at different timestamps $t$ and $t'$ during a multi-step attack have very different magnitude, varying from $\sim0$ to $\geq 100$ for CIFAR-100. 
To take full advantage of the intermediate-level discrepancies in Eq.~\eqref{eq:opt0}, we suggest performing data normalization before solving the linear regression problem. That being said, we suggest $\tilde{\mathbf w}^\ast = (\tilde{\mathbf H}^T \tilde{\mathbf H} + \lambda \mathbf I_m)^{-1} \tilde{\mathbf H}^T \mathbf r$, in which the $t$-th row of the matrix $\tilde{\mathbf H}$ is the normalized intermediate-level discrepancy $(\mathbf h^{\mathrm{adv}}_t-\mathbf h^{\mathrm{adv}}_0)/\|\mathbf h^{\mathrm{adv}}_t-\mathbf h^{\mathrm{adv}}_0\|$ obtained at the $t$-th iteration of the baseline attack. We here optimize a similar problem as in Eq.~\eqref{eq:opt0}, i.e., 
\begin{equation}
    \max_{\mathbf{\Delta}_x}\ (g(\mathbf x+\mathbf{\Delta}_x) - \mathbf h^{\mathrm{adv}}_0)^T \tilde{\mathbf w}^\ast, \quad \mathrm{s.t.}\ (\mathbf x+\mathbf{\Delta}_x) \in \Psi,
\end{equation}
as both $\frac{(g(\mathbf x+\mathbf{\Delta}_x) - \mathbf h^{\mathrm{adv}}_0)^T \tilde{\mathbf w}^\ast}{\|g(\mathbf x+\mathbf{\Delta}_x) - \mathbf h^{\mathrm{adv}}_0\|}$ and $\|g(\mathbf x+\mathbf{\Delta}_x) - \mathbf h^{\mathrm{adv}}_0\|$ are expected to be maximized. 

\section{Experimental Results}

In this section, we show experimental results to verify the efficacy of our method. We will first compare the usefulness of different intermediate-level discrepancies when being applied as the directional guides in our framework and ILA, and then compare plausible settings of our method on CIFAR-100. 
We will show that our method significantly outperforms its competitors on CIFAR-100 and ImageNet in Section~\ref{sec:sota}.
Our experimental setting are deferred to Section~\ref{sec:setting}.

\begin{figure}[t]
	\centering 
	\subfigure[]{\label{fig:1a}\includegraphics[width=0.493\linewidth]{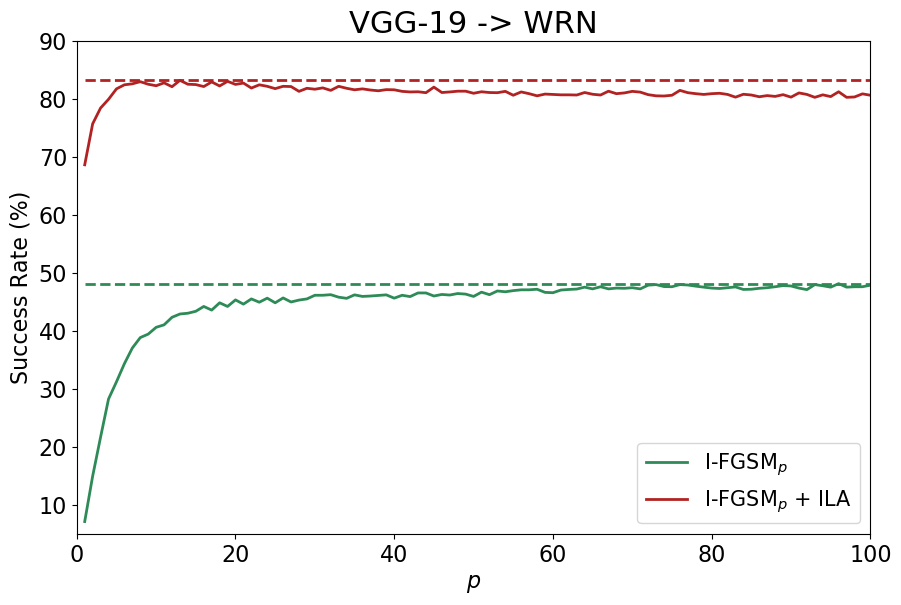}}
	\subfigure[]{\label{fig:1b}\includegraphics[width=0.493\linewidth]{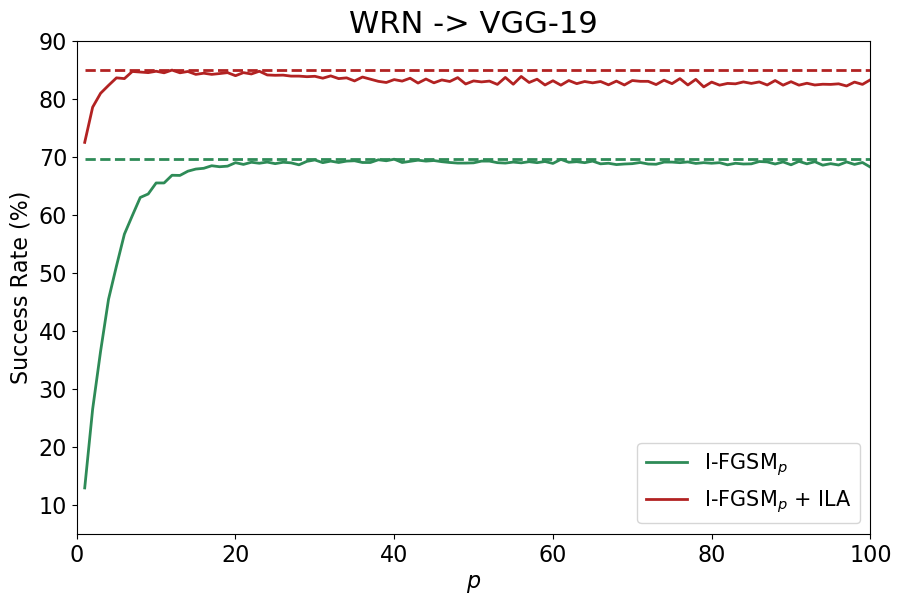}}\vskip -0.2in
	\caption{How the transferability of the baseline adversarial example (a) crafted on VGG-19 to attack WRN (enhanced by ILA or not) and (b) crafted on WRN to attack VGG-19 (enhanced by ILA or not) varies with $p$. The dashed lines indicate the performance with the optimal $p$ values. We see that the most transferable I-FGSM$_p$+ILA examples ($\epsilon=0.03$) are obtained around $p=10$, and the success rate declines consistently with greater $p$ for $p\geq 10$.} \vskip -0.25in
\label{fig:1}
\end{figure}

\subsection{Delve into the Multi-step Baseline Attacks}

We conducted a comprehensive study on the adversarial transferability of contemporary results in multi-step baseline attacks and how competent they are in assisting subsequent methods like ILA~\cite{Huang2019} and ours. We performed experiments on CIFAR-100~\cite{Krizhevsky2009}, an image classification dataset that consisting of \num{60000} images from \num{100} classes. It was officially divided into a training set of \num{50000} images and a test set of \num{10000} images. We considered two models in this study: VGG-19~\cite{Simonyan2015} with batch normalization~\cite{Ioffe2015} and wide ResNet (WRN)~\cite{Zagoruyko2016}, (specifically, WRN-28-10). Their architectures are very different, since the latter is equipped with skip connections and it is much deeper than the former. We collected pre-trained models from Github~\footnote{https://github.com/bearpaw/pytorch-classification}, and they show 28.05\% and 18.14\% prediction errors respectively on the official test set. We randomly chose \num{3000} images that could be correctly classified by the two models to initialize the baseline attack, and the success rate over \num{3000} crafted adversarial examples was considered. 

We applied I-FGSM as the baseline attack and utilized adversarial examples crafted on one model (i.e., VGG-19/WRN) to attack the other (i.e., WRN/VGG-19).
We tested the success rate when: (1) directly adopting the generated I-FGSM adversarial examples on the victim models and (2) adopting ILA on the basis of I-FGSM. 
Untargeted attacks were performed under a constraint of the $\ell_\infty$ norm with $\epsilon=0.03$.
We denote by I-FGSM$_p$ the results of I-FGSM running for $p$ steps, and denote by I-FGSM$_p$+ILA the ILA outcomes on the basis of I-FGSM$_p$.
The success rates of using one model to attack the other are summarized in Fig.~\ref{fig:1}, with varying $p$.
Apparently, ILA operates better with relatively earlier results from I-FGSM (i.e., I-FGSM$_p$ with a relatively smaller $p$).
The most transferable adversarial examples can be gathered around $p=10$ when it is equipped with ILA, and further increasing $p$ would lead to declined success rates. While \emph{without ILA}, running more I-FGSM iterations are more beneficial to the transferability.

\begin{figure}[t]
	\centering 
	\subfigure[]{\label{fig:2a}\includegraphics[width=0.495\textwidth]{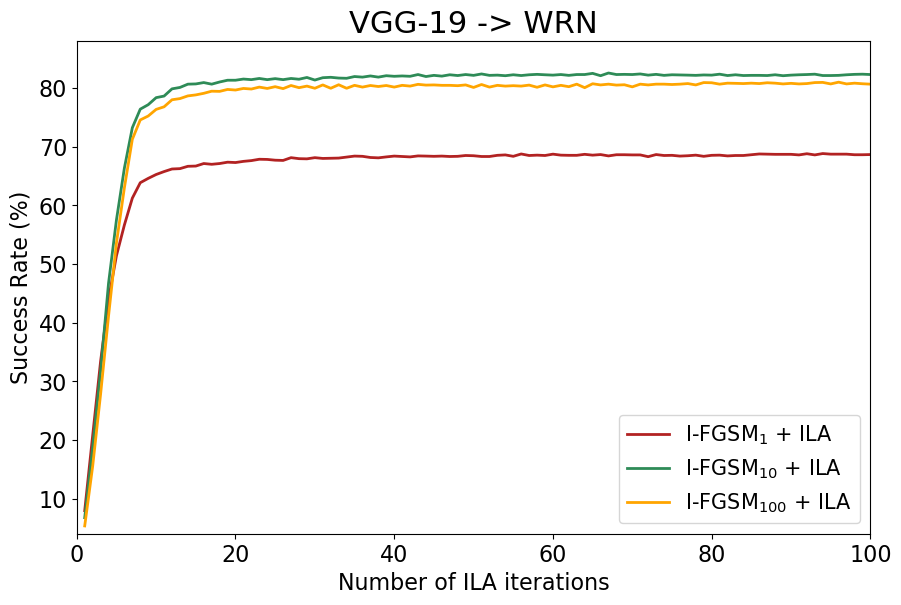}}
	\subfigure[]{\label{fig:2b}\includegraphics[width=0.495\textwidth]{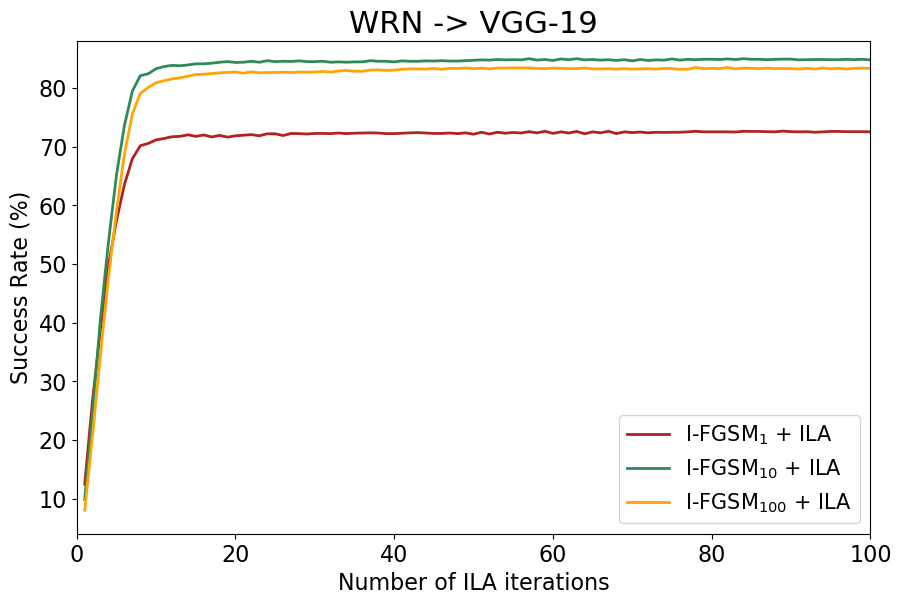}}\vskip -0.15in
	\caption{How the transferability of I-FGSM adversarial examples, (a) crafted on VGG-19 to attack WRN and (b) crafted on WRN to attack VGG-19, are enhanced by \emph{ILA}. We let $\epsilon=0.03$.} \vskip -0.2in
\label{fig:2}
\end{figure}

\begin{figure}[ht!]
	\centering \vskip -0.2in
	\subfigure[]{\label{fig:3a}\includegraphics[width=0.495\textwidth]{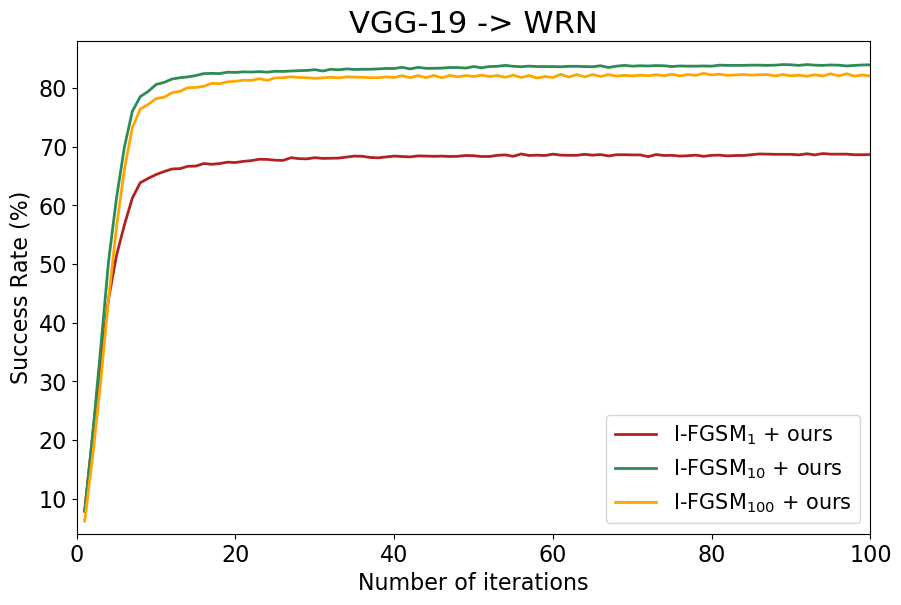}}
	\subfigure[]{\label{fig:3b}\includegraphics[width=0.495\textwidth]{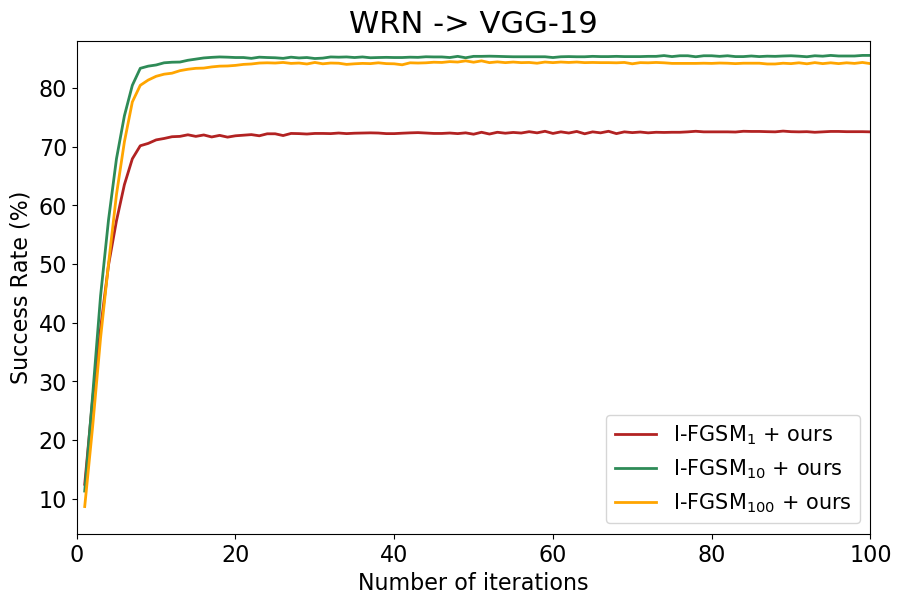}}\vskip -0.15in
	\caption{How the transferability of I-FGSM examples, (a) crafted on VGG-19 to attack WRN and (b) crafted on WRN to attack VGG-19, are enhanced by \emph{our method}. The range of the $y$ axes are kept the same as in Fig.~\ref{fig:2} for easy comparison. We let $\epsilon=0.03$.}  \vskip -0.2in
\label{fig:3}
\end{figure}

\begin{figure}[t]
	\centering \vskip -0.2in
	\subfigure[]{\label{fig:4a}\includegraphics[width=0.495\textwidth]{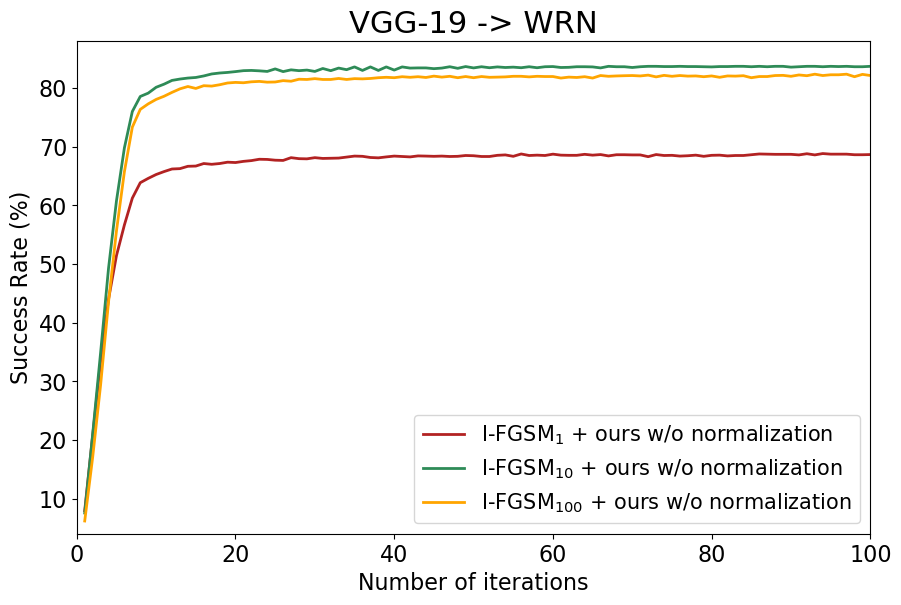}}
	\subfigure[]{\label{fig:4b}\includegraphics[width=0.495\textwidth]{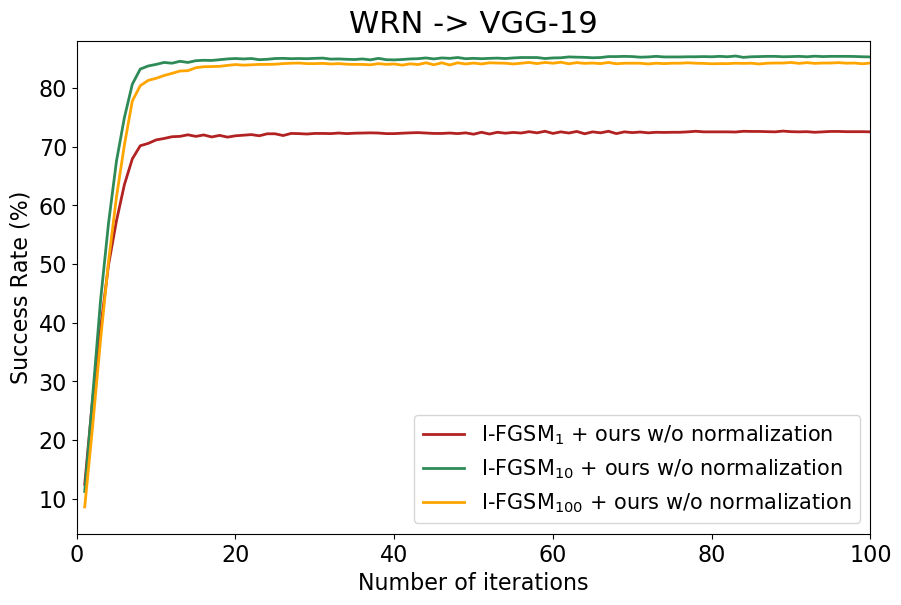}}\vskip -0.15in
	\caption{How the transferability of I-FGSM examples, (a) crafted on VGG-19 to attack WRN and (b) crafted on WRN to attack VGG-19, are enhanced by \emph{our method}. Intermediate-level normalization is \emph{NOT} performed. The range of the $y$ axes are kept the same as with Fig.~\ref{fig:2} and~\ref{fig:3} for easy comparison. We let $\epsilon=0.03$.} \vskip -0.1in
\label{fig:4}
\end{figure}

In more detail, Fig.~\ref{fig:2} shows how much the transferability is improved along with ILA. We see that I-FGSM$_{10}$+ILA consistently outperforms I-FGSM$_{100}$+ILA.
We evaluated the performance of our method based on I-FGSM$_p$ examples similarly, and the results are illustrated in Fig.~\ref{fig:3} and~\ref{fig:4}, one with intermediate-level normalization and the other without.
We set $\lambda\rightarrow\infty$, and how the performance of our method varies with $\lambda$ will be discussed in Section~\ref{sec:lambda}.
Obviously, the same tendency as demonstrated in Fig.~\ref{fig:2} can also be observed in Fig.~\ref{fig:3} and~\ref{fig:4}.
That being said, earlier results from the multi-step baseline attack I-FGSM are more effective as guide directions for both ILA and our method. As illustrated in Fig.~\ref{fig:1}, the baseline attack converges faster than expected, making many ``training samples'' in $\{(\mathbf h_t^{\mathrm{adv}}-\mathbf h^{\mathrm{adv}}_0, l_t)\}$ highly correlated, with or without intermediate-level normalization. 
Using relatively early results relieve the problem and is thus beneficial to our method. The performance gain on ILA further suggests that earlier results from I-FGSM overfit less on the source model, and they are more suitable as the directional guides.
In what follows, we fix $p=10$ without any further clarification, which also reduces the computational complexity of our method for calculating $\mathbf w^\ast$ or $\tilde{\mathbf w}^\ast$ (by at least $10\times$), in comparison with $p=100$. 

Notice that the ranges of the vertical axes are the same for Fig.~\ref{fig:2},~\ref{fig:3}, and~\ref{fig:4}. It can easily be observed from the figures that our method, with either $p=10$ or $p=100$, achieves superior performance in comparison with ILA in the same setting.
With $p=1$, the two methods demonstrate exactly the same results, as has been discussed in Section~\ref{sec:method}.
More comparative studies will be conducted in Section~\ref{sec:sota}.
Based on I-FGSM$_{10}$, our method shows a success rate of $85.53\%$ with normalization and $85.27\%$ without, when attacking VGG-19 using WRN.
That being said, the intermediate-level normalization slightly improves our method, and we will keep it for all the experiments in the sequel of the paper.

\subsection{Our Method with Varying $\lambda$}\label{sec:lambda}

\begin{figure}[t]
	\centering 
	\includegraphics[width=0.63\textwidth]{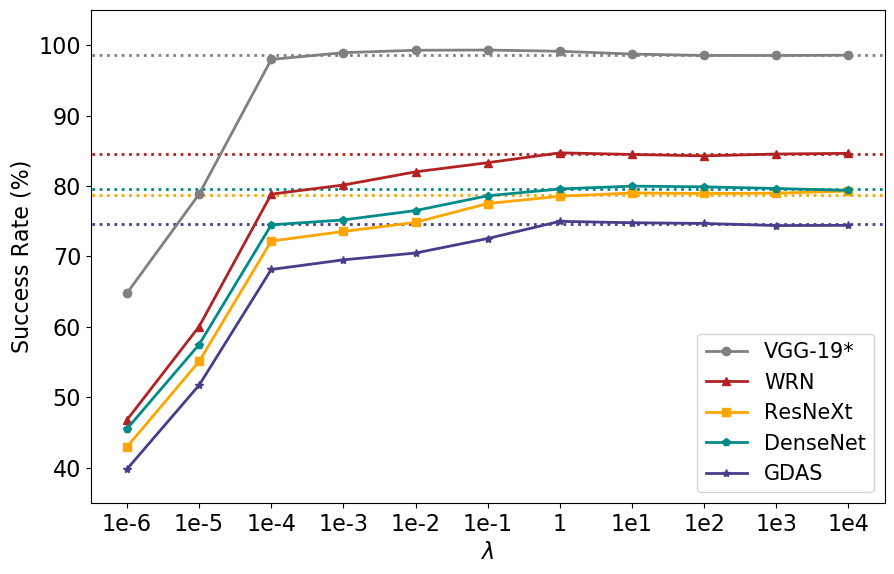} \vskip -0.15in
	\caption{How the performance of our method varies with $\lambda$. The dotted lines indicates the success rates when setting $\lambda\rightarrow \infty$. We let $\epsilon=0.03$.} \vskip -0.2in
\label{fig:lambda}
\end{figure}

One seemingly important hyper-parameter in our method is $\lambda$, which controls the smoothness of the linear regression model with $\mathbf w^\ast$ or $\tilde{\mathbf w}^\ast$. 
Here we report the performance with varying $\lambda$ values and evaluate how different choices for $\lambda$ affects the final result. The experiment was also performed on CIFAR-100. 
To make the study more comprehensive, we tested with a few more victim models, including a ResNeXt, a DenseNet, and a convolutional network called GDAS~\cite{Dong2019}~\footnote{https://github.com/D-X-Y/AutoDL-Projects} whose architecture is learned via neural architecture search~\cite{Zoph2016neural,Zela2020bench}. 
We used VGG-19 as the source model and the others (i.e., WRN, ResNeXt, DenseNet, and GDAS) as victim models.
Obviously in Fig.~\ref{fig:lambda}, small $\lambda$ values lead to unsatisfactory success rates on the source and victim models, and relatively large $\lambda$s (even approaching infinity) share similar performance. Specifically, the optimal average success rate $79.55\%$ is obtained with $\lambda=10$, while we can still get $79.38\%$ with $\lambda \rightarrow \infty$. Here we would like to mention that, with the Woodbury identity and a scaling factor $1/\lambda$ in Eq.~\eqref{eq:woodbury} being eliminated when it is substituted into~\eqref{eq:opt1}, we have $\mathbf H^T(\mathbf H \mathbf H^T + \lambda\mathbf I)^{-1} \mathbf H\rightarrow \mathbf 0$, \emph{but NOT} $\mathbf w^\ast \rightarrow \mathbf 0$ or $\tilde{\mathbf w}^\ast \rightarrow \mathbf 0$. 

As has been discussed in Section~\ref{sec:method}, setting an infinitely large $\lambda$ leads to a simpler optimization problem and lower computational cost when computing the parameters of the linear mapping. In particular, the calculation of matrix inverse can be omitted with $\lambda \rightarrow \infty$. Although not empirically optimal, letting $\lambda \rightarrow \infty$ leads to decent performance of our method, hence in the following experiments we fixed the setting and solved the optimization problem in Eq.~\eqref{eq:opt1} instead of~\eqref{eq:opt0} to keep the introduced computational burden on the intermediate layer at a lower level. As such, the main computational cost shall come from back-propagation, which is inevitable in ILA as well.
We compare the run-time of our method with that of ILA on both CPU and GPU in Table~\ref{tab:runtime}, where how long it takes to craft 100 adversarial examples on the VGG-19 source model by using the two methods is reported. The experiment was performed on an Intel Xeon Platinum CPU and an NVIDIA Tesla-V100 GPU. The code was implemented using PyTorch~\cite{Paszke2019pytorch}. For fair comparison, the two methods were both executed for $100$ iterations for generating a single adversarial example. 
It can be seen that both methods show similar run-time in practice.

\begin{table}[ht] \vskip -0.15in
\caption{Run-time comparison between our method and ILA.}\label{tab:runtime}
\centering
    \begin{tabular}{C{0.65in}C{0.67in}C{0.67in}}
    \hline
    & CPU (s) & GPU (s)\\
    \hline
    ILA~\cite{Huang2019}  &70.073 &2.336  \\
    Ours &71.708 &2.340  \\
    \hline
    \end{tabular}\vskip -0.25in
\end{table}

\subsection{Compare with The State-of-the-arts}\label{sec:sota}

In this subsection, we show more experimental results to verify the effectiveness of our method. We tried attacking 14 models on CIFAR-100 and ImageNet, 4 for the former and 10 for the latter. For CIFAR-100, the victim models include WRN, ResNeXt, DenseNet, and GDAS, as previously introduced, while for ImageNet, the tested victim models include VGG-19~\cite{Simonyan2015} with batch normalization, ResNet-152~\cite{He2016}, Inception v3~\cite{Szegedy2016}, DenseNet~\cite{Huang2017densely}, MobileNet v2~\cite{Sandler2018mobilenetv2}, SENet~\cite{Hu2018}, ResNeXt~\cite{Xie2017aggregated} (precisely ResNeXt-101-32x8d), WRN~\cite{Zagoruyko2016} (precisely WRN-101-2), PNASNet~\cite{Liu2018}, and MNASNet~\cite{Tan2019mnasnet}. 
The source models for the two datasets are VGG-19 and ResNet-50, respectively.
Pre-trained models on ImageNet are collected from Github\footnote{https://github.com/Cadene/pretrained-models.pytorch} and the torchvision repository\footnote{https://github.com/pytorch/vision/tree/master/torchvision/models}.
We mostly compare our method with ILA, since the two share the same problem setting as introduced in Section~\ref{sec:problem}. We first compare their performance on the basis of I-FGSM, which is viewed as the most basic baseline attack in the paper. 
Table~\ref{tab:cifar} and~\ref{tab:imagenet} summarize the results on CIFAR-100 and ImageNet, respectively.
It can be seen that our method outperforms ILA remarkably on almost all test cases.
As has been explained, both methods work better with relatively earlier I-FGSM results. The results in Table~\ref{tab:cifar} and~\ref{tab:imagenet} are obtained with $p=10$. We also tested with $p=100$, $30$, and $20$, and our method is superior to ILA in all these settings. Both methods chose the same intermediate layer according to the procedure introduced in~\cite{Huang2019}.

\begin{table}[t!]
 \caption{Performance of transfer-based attacks on CIFAR-100 using I-FGSM with an $\ell_\infty$ constraint of the adversarial perturbation in the untargeted setting. The symbol * indicates when the source model is the target. The best average results are in red. }
 \vskip -0.2in
 \label{tab:cifar}
 \begin{center}\resizebox{0.99\linewidth}{!}{
   \begin{tabular}{C{0.65in}C{0.67in}C{0.48in}C{0.62in}C{0.62in}C{0.62in}C{0.62in}C{0.62in}C{0.62in}}

    \hline
    Dataset                    & Method  & $\epsilon$ & VGG-19* & WRN  & ResNeXt  & DenseNet  & GDAS & \textbf{Average}  \\
    \hline
    \multirow{10}{*}{CIFAR-100} & \multirow{3}{*}{-} & 0.1 & 100.00\% & 74.90\% & 69.33\% & 71.77\% & 66.93\% & 70.73\%   \\
                                                  & & 0.05 & 100.00\% & 64.67\% & 57.63\% & 61.13\% & 56.00\% & 59.86\%         \\
                                                  & & 0.03 & 100.00\% & 48.27\% & 41.20\% & 43.83\% & 39.13\% & 43.11\%      \\
    \cmidrule(r){2-9}
                               & \multirow{3}{*}{ILA~\cite{Huang2019}} & 0.1 & 99.07\% & 97.53\% & 96.90\% & 97.30\% & 96.03\% & 96.94\%         \\
                                                   & & 0.05 & 99.03\% & 93.90\% & 90.73\% & 91.60\% & 88.73\% & 91.24\%         \\
                                                   & & 0.03 & 98.77\% & 82.73\% & 76.53\% & 77.87\% & 72.83\% & 77.49\%      \\
    \cmidrule(r){2-9}
                               & \multirow{3}{*}{Ours} & 0.1  & 98.83\% & 97.80\% & 97.07\% & 97.50\% & 96.51\% & \textcolor{red}{\textbf{97.22\%}}         \\
                                                   & & 0.05 & 98.87\% & 94.03\% & 91.27\% & 91.73\% & 89.37\% & \textcolor{red}{\textbf{91.60\%}}        \\
                                                   & & 0.03 & 98.53\% & 84.57\% & 78.70\% & 79.60\% & 74.63\% & \textcolor{red}{\textbf{79.38\%}}      \\
    \hline
   \end{tabular}}
 \end{center}\vskip -0.2in
\end{table} 

\begin{table}[t!]
 \caption{Performance of transfer-based attacks on ImageNet using I-FGSM with $\ell_\infty$ constraint in the untargeted setting. We use the symbol * to indicate when the source model is used as the target. The lower sub-table is the continuation of the upper sub-table. The best average results are marked in red. }\vskip -0.2in
 \label{tab:imagenet}
 \begin{center}\resizebox{0.99\linewidth}{!}{
   \begin{tabular}{C{0.64in}C{0.62in}C{0.42in}C{0.7in}C{0.7in}C{0.7in}C{0.7in}C{0.7in}C{0.74in}}

    \hline
    Dataset                    & Method  & $\epsilon$ & ResNet-50* & VGG-19  & ResNet-152$\,$  & Inception$\,$v3  & DenseNet & MobileNet$\,$v2  \\
    \hline
    \multirow{10}{*}{ImageNet} & \multirow{3}{*}{-} & 0.1  & 100.00\% & 67.70\% & 61.10\% & 36.36\% & 65.00\% & 65.60\% \\
 &  & 0.05 & 100.00\% & 54.46\% & 44.74\% & 24.68\% & 49.90\% & 52.12\% \\
 &  & 0.03 & 100.00\% & 36.80\% & 26.56\% & 13.72\% & 32.08\% & 34.56\% \\
    \cmidrule(r){2-9}
                              & \multirow{3}{*}{ILA~\cite{Huang2019}} & 0.1  & 99.96\%  & 97.62\% & 96.96\% & 87.94\% & 96.76\% & 96.54\% \\
 &  & 0.05 & 99.96\%  & 88.74\% & 86.02\% & 61.20\% & 86.42\% & 85.62\% \\
 &  & 0.03 & 99.96\%  & 69.96\% & 63.14\% & 34.86\% & 64.52\% & 65.68\% \\
    \cmidrule(r){2-9}
                              & \multirow{3}{*}{Ours} & 0.1  & 99.92\%  & 97.60\% & 96.98\% & 88.46\% & 97.02\% & 96.74\% \\
 &  & 0.05 & 99.92\%  & 89.40\% & 87.12\% & 64.96\% & 88.14\% & 86.98\% \\
 &  & 0.03 & 99.90\%  & 72.88\% & 67.82\% & 39.40\% & 68.38\% & 69.20\% \\
  \hline
\\
    \hline
    Dataset  & Method  & $\epsilon$ & SENet & ResNeXt  & WRN  & PNASNet  & MNASNet & \textbf{Average}  \\
    \hline
    \multirow{10}{*}{ImageNet} & \multirow{3}{*}{-} & 0.1  & 45.32\%  & 56.36\% & 56.96\% & 35.34\% & 63.68\% & 55.34\% \\
 &  & 0.05 & 29.92\%  & 41.74\% & 40.82\% & 22.76\% & 49.46\% & 41.06\% \\
 &  & 0.03 & 15.94\%  & 23.46\% & 24.32\% & 11.90\% & 33.12\% & 25.25\% \\
    \cmidrule(r){2-9}
                              & \multirow{3}{*}{ILA~\cite{Huang2019}} & 0.1  & 93.76\%  & 96.00\% & 95.62\% & 91.04\% & 96.70\% & 94.89\% \\
 &  & 0.05 & 74.36\%  & 82.54\% & 81.80\% & 65.74\% & 84.32\% & 79.68\% \\
 &  & 0.03 & 46.50\%  & 59.24\% & 58.58\% & 37.22\% & 64.78\% & 56.45\% \\
    \cmidrule(r){2-9}
                              & \multirow{3}{*}{Ours} & 0.1  & 94.00\%  & 96.16\% & 95.74\% & 91.22\% & 96.86\% & \color{red} {\textbf{95.08\%}} \\
 &  & 0.05 & 76.26\%  & 84.00\% & 83.50\% & 69.24\% & 86.26\% & {\color{red} \textbf{81.59\%}} \\
 &  & 0.03 & 50.26\%  & 63.48\% & 62.72\% & 42.16\% & 67.94\% & {\color{red} \textbf{60.42\%}} \\
  \hline
   \end{tabular}   
   }
 \end{center}\vskip -0.32in
\end{table} 

In addition to ILA, there exist several other methods in favor of black-box transferability, yet most of them are orthogonal to our method and ILA and they can be applied as baseline attacks in a similar spirit to the I-FGSM baseline.
We tried adopting the two methods based on PGD~\cite{Madry2018}, TAP~\cite{Zhou2018}, and I-FGSM with momentum (MI-FGSM)~\cite{Dong2018}, which are probably more powerful than I-FGSM. Their default setting all choose the cross-entropy loss for optimization with an $\ell_\infty$ constraint. In fact, I-FGSM can be regarded as a special case of the PGD attack with a random restart radius of zero. TAP and MI-FGSM are specifically designed for transfer-based black-box attacks and the generated adversarial examples generally show better transferability than the I-FGSM examples. Table~\ref{tab:attacks} shows that TAP outperforms the other three (including the basic I-FGSM) multi-step baselines. MI-FGSM and PGD are the second and third best, while the basic I-FGSM performs the worst in the context of adversarial transferability \emph{without further enhancement}.
Nevertheless, when further equipped with our method or ILA for transferability enhancement, PGD and I-FGSM become the second and third best, respectively, and TAP is still the winning solution showing $64.21\%$ success rates.
The MI-FGSM-related results imply that introducing momentum leads to less severe overfitting on the source model, yet such a benefit diminishes when being used as directional guides for ILA and our method.
Whatever baseline attack is applied, our method always outperforms ILA in our experiment, which is conducted on ImageNet with $\epsilon=0.03$.

\begin{table}[t!]
 \caption{Performance of transfer-based attacks on ImageNet. Different baseline attacks are compared in the same setting of $\epsilon=0.03$. The best average result is marked in red. }\vskip -0.2in
 \label{tab:attacks}
 \begin{center}\resizebox{0.99\linewidth}{!}{
   \begin{tabular}{L{0.8in}C{0.47in}C{0.45in}C{0.45in}C{0.47in}C{0.45in}C{0.45in}C{0.47in}C{0.45in}C{0.46in}}

    \hline
    & \multicolumn{3}{c}{MI-FGSM~\cite{Dong2018}}  & \multicolumn{3}{c}{PGD~\cite{Madry2018}} &  \multicolumn{3}{c}{TAP~\cite{Zhou2018}} \\
    \cmidrule(r){2-4}\cmidrule(r){5-7}\cmidrule(r){8-10}
    & - & ILA~\cite{Huang2019} & Ours & - & ILA~\cite{Huang2019} & Ours & - & ILA~\cite{Huang2019} & Ours  \\
    \hline
        ResNet-50*        & 100.00\% & 99.94\% & 99.90\%   & 100.00\% & 99.94\% & 99.88\%   & 100.00\% & 99.98\% & 99.96\%   \\
        VGG-19           & 46.46\%  & 67.18\% & 70.28\%   & 40.80\%  & 70.38\% & 72.22\%   & 58.34\%  & 78.00\% & 77.96\%   \\
        ResNet-152       & 37.90\%  & 60.76\% & 63.62\%   & 31.06\%  & 64.32\% & 68.02\%   & 45.04\%  & 67.42\% & 68.52\%   \\
        Inception v3     & 21.50\%  & 33.98\% & 37.26\%   & 16.60\%  & 37.76\% & 41.52\%   & 25.50\%  & 40.70\% & 42.88\%   \\
        DenseNet         & 42.14\%  & 63.02\% & 65.86\%   & 37.78\%  & 67.14\% & 69.94\%   & 49.02\%  & 70.56\% & 71.98\%   \\
        MobileNet v2     & 45.78\%  & 63.92\% & 67.04\%   & 39.02\%  & 66.62\% & 69.66\%   & 54.98\%  & 72.72\% & 73.84\%   \\
        SENet            & 24.60\%  & 45.26\% & 48.14\%   & 18.28\%  & 46.32\% & 49.60\%   & 33.68\%  & 55.30\% & 56.26\%   \\
        ResNeXt          & 34.28\%  & 56.08\% & 59.64\%   & 27.78\%  & 60.16\% & 63.72\%   & 41.30\%  & 64.50\% & 66.20\%   \\
        WRN              & 34.20\%  & 56.28\% & 59.66\%   & 27.92\%  & 60.08\% & 62.82\%   & 45.08\%  & 66.24\% & 67.06\%   \\
        PNASNet          & 18.36\%  & 34.82\% & 38.56\%   & 13.82\%  & 38.50\% & 42.68\%   & 22.20\%  & 42.24\% & 44.76\%   \\
        MNASNet          & 43.26\%  & 62.34\% & 65.36\%   & 37.08\%  & 65.08\% & 67.76\%   & 53.64\%  & 71.64\% & 72.64\%   \\
        \textbf{Average} & 34.85\%  & 54.36\% & 57.54\%   & 29.01\%  & 57.64\% & 60.79\%   & 42.88\%  & 62.93\% & {\color{red} \textbf{64.21\%}}   \\
    \hline
   \end{tabular}}
 \end{center}\vskip -0.32in
\end{table} 

\begin{figure}[ht!]
	\centering \vskip -0.25in
	\subfigure[]{\label{fig:6a}\includegraphics[width=0.495\textwidth]{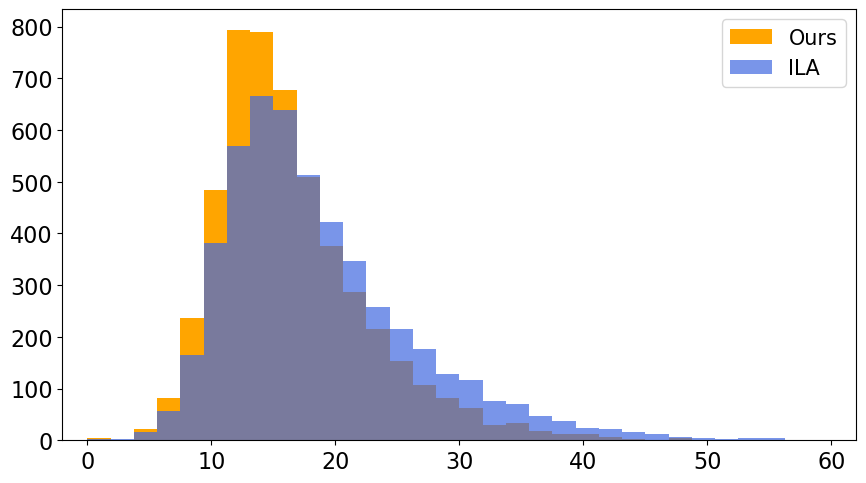}}
	\subfigure[]{\label{fig:6b}\includegraphics[width=0.495\textwidth]{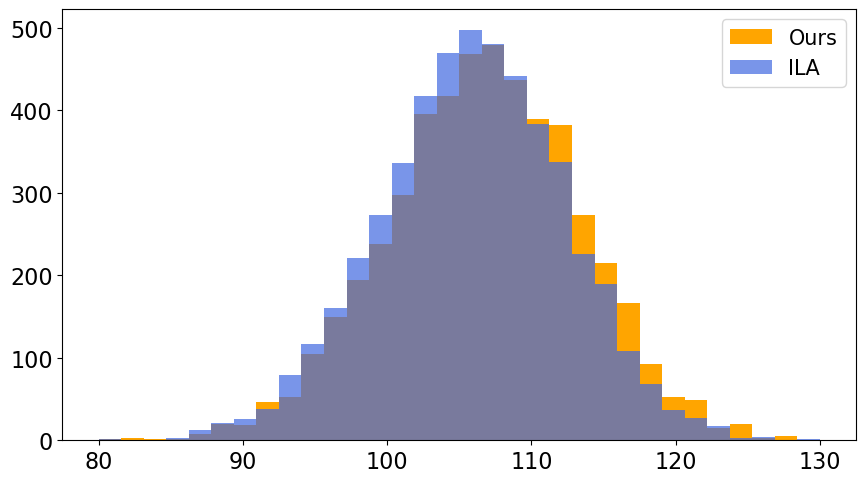}}\vskip -0.15in
	\caption{Comparison of our method and ILA in the sense of (a) the cross-entropy loss and (b) intermediate-level disturbance on the ResNet-50 source model on ImageNet.} \vskip -0.2in
\label{fig:analysis}
\end{figure}

It can be observed from all results thus far that our method bears a slightly decreased success rate on the source model, yet it delivers an increased capability of generating transferable adversarial examples. 
It is discussed in Section~\ref{sec:method} that our method provides an advantage over the status quo that it is not guaranteed to achieve optimal intermediate-level disturbance.
To further analyze the functionality of our method, we illustrate the cross-entropy loss and intermediate-level disturbance on the ImageNet adversarial examples crafted using our method and ILA in Fig.~\ref{fig:analysis}. It depicts that our method gives rise to larger intermediate-level disturbance in comparison to ILA, with a little sacrifice of the adversarial loss. As has been explained, more significant intermediate-level disturbance indicates higher transferability in general, which well-explains the superiority of our method in practice.
Fig.~\ref{fig:analysis} also demonstrates the slightly deteriorating effect on the performance of our method in the white-box setting, which does not really matter under the considered threat model though.

The success of our method is ascribed to effective aggregations of diverse directional guides over the whole procedure of the given baseline attack. Obviously, it seems also plausible to ensemble different baseline attacks to gain even better results in practice, since this probably gives rise to a more informative and sufficient ``training set'' for predicting the adversarial loss linearly. To experimentally testify the conjecture, we directly collected $20$ intermediate-level discrepancies and their corresponding adversarial loss from two of the introduced baseline attacks: I-FGSM and PGD, for learning the linear regression model, i.e., $10$ from each of them, and we tested our method similarly. We evaluated the performance of such an straightforward ensemble on ImageNet and it shows an average success rate of $62.82\%$ under $\epsilon=0.03$. Apparently, it is superior to that using the I-FGSM ($60.42\%$) or PGD ($60.79\%$) baseline results solely.

\begin{figure}[ht]
	\centering \vskip -0.25in
	\includegraphics[width=0.57\textwidth]{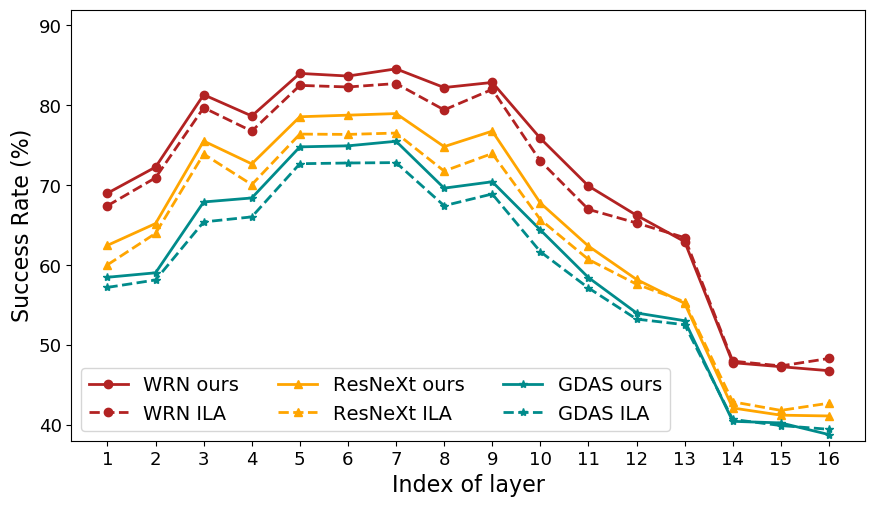}\vskip -0.18in
	\caption{Comparison of our method and ILA on CIFAR-100 with varying choices of the intermediate layer (on the VGG-19 source model) to calculate the intermediate-level discrepancies. Best viewed in color. We tested under $\epsilon=0.03$.} \vskip -0.2in
\label{fig:location_cifar}
\end{figure}

\begin{figure}[ht]
	\centering \vskip -0.25in
	\includegraphics[width=0.57\textwidth]{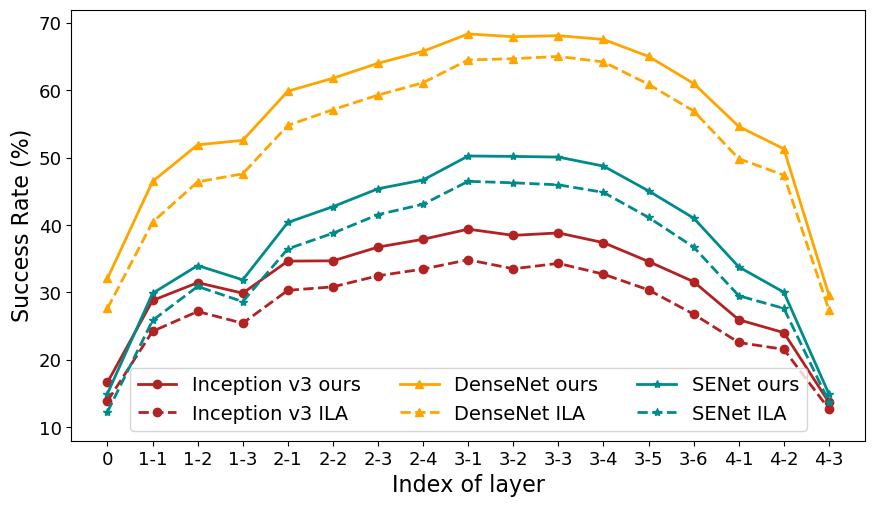}\vskip -0.18in
	\caption{Comparison of our method and ILA on ImageNet with varying choices of the intermediate layer (on the ResNet-50 source model) to calculate the intermediate-level discrepancies. The index of layer shown as ``3-1'' indicates the first block of the third meta-block. Best viewed in color. We tested under $\epsilon=0.03$.} \vskip -0.28in
\label{fig:location_imagenet}
\end{figure}

Since $\lambda\rightarrow \infty$ was set, our method did not fine-tune more hyper-parameters compared to ILA. A crucial common hyper-parameter of the two methods is the location where the intermediate-level discrepancies are calculated. We compare them with various settings of the location on CIFAR-100 in Fig.~\ref{fig:location_cifar}. It can be seen that our method consistently outperforms ILA in almost all test cases from the first to the 13-th layer on the source model VGG-19. Both methods achieve their optimal results \emph{at the same location}, therefore we can use the same procedure for selecting layers as introduced in ILA. The results on DenseNet are very similar to those on ResNeXt, and thus not plotted for clearer illustration. 
Notice that even the worst results of the intermediate-level methods on these victim models are better than the baseline results.
The ImageNet results on three representative victim models 
are given in Fig.~\ref{fig:location_imagenet}, and the same conclusions can be made. For CIFAR-100, the intermediate-level discrepancies were calculated right after each convolutional layer, while for ImageNet, we calculated at the end of each computational block.

\subsection{Experimental Settings and $\ell_2$ attacks}\label{sec:setting}

We mostly consider untargeted $\ell_\infty$ attacks in the black-box setting, just like prior arts of our particular interest~\cite{Huang2019,Inkawhich2019,Zhou2018}. The element-wise maximum allowed perturbation, i.e., the $\ell_\infty$ norm, was constrained to be lower than a positive scalar $\epsilon$. We tested with $\epsilon=0.1$, $0.05$, and $0.03$ in our experiments. 
In addition to the $\ell_\infty$ attacks, $\ell_2$ attacks were also tested and the same conclusions could be made, i.e., our method still outperforms ILA and the original baseline considerably.
Owing to the space limit of the paper, we only report some representative results here.
On CIFAR-100, the I-FGSM baseline achieves an average success rate of $47.23\%$, based on which ILA and our method achieve $73.73\%$ and $75.78\%$, respectively. To be more specific, on the victim models including VGG-19*, our method achieves $81.27\%$ (for WRN), $75.4\%$ (for ResNeXt), $75.60\%$ (for DenseNet), $70.83\%$ (for GDAS), and $97.73\%$ (for VGG-19*) success rates.
On ImageNet, our method is also remarkably superior to the two competitors in the sense of the average $\ell_2$ success rate (ours: $76.73\%$, ILA: $74.68\%$, and the baseline: $54.79\%$). The same victim models as in Table~\ref{tab:imagenet} were used. On CIFAR-100 and ImageNet, the $\ell_2$ norm of the perturbations was constrained to be lower than $1.0$ and $10$, respectively.

In $\ell_\infty$ cases, the step-size for I-FGSM, PGD, TAP, and MI-FGSM were uniformly set as $1/255$, on both CIFAR-100 and ImageNet, while in $\ell_2$ cases, we used $0.1$ and $1.0$ on the two datasets respectively. Other hyper-parameters were kept the same for all methods under both the $\ell_\infty$ and $\ell_2$ constraints.
We randomly sampled \num{3000} and \num{5000} test images that are correctly classified by the victim models from the two datasets respectively to initialize the baseline attacks and generate \num{3000} and \num{5000} adversarial examples using each method, as suggested in many previous works in the literature. For CIFAR-100, they were sampled from the official test set consisting of \num{10000} images, while for ImageNet, they were sampled from the validation set. We run our method and ILA for \num{100} iterations on the two datasets, such that they \emph{both reached performance plateaux}. Input images to all DNNs were re-scaled to $[0,1]$ and the default pre-processing pipeline was adopted when feeding images to the DNNs. For ILA and TAP, followed the open-source implementation from Huang et al.~\cite{Huang2019}. 
An optional setting for implementing transfer-based attacks is to save the adversarial examples as bitmap images (or not) before feeding them to the victim models. The adversarial examples will be in an 8-bit 
image format for the former and a 32-bit floating-point format for the latter. We consider the former to be more realistic in practice. 

Our learning objective does not employ an \emph{explicit} term for encouraging large norms of the intermediate-level discrepancies. It is possible to further incorporate one such term in~\eqref{eq:opt0} and~\eqref{eq:opt1}. However, an additional hyper-parameter will be introduced inevitably, as discussed by ILA regarding the flexible loss~\cite{Huang2019}. We shall consider such a formulation in future work. 
Our code is at https://github.com/\\qizhangli/ila-plus-plus.

\section{Conclusions} 

In this paper, we have proposed a novel method for improving the transferability of adversarial examples. It operates on baseline attack(s) whose optimization procedures can be analyzed to extract a set of directional guides. By establishing a linear mapping to estimating the adversarial loss using intermediate-layer feature maps, we have developed an adversarial objective function that could take full advantage of the baseline attack(s). The effectiveness of our method has been shown via comprehensive experimental studies on CIFAR-100 and ImageNet. 


  
\paragraph{\textbf{Acknowledgment.}} This material is based upon work supported by the National Science Foundation under Grant No.\ 1801751.

This research was partially sponsored by the Combat Capabilities Development Command Army Research Laboratory and was accomplished under Cooperative Agreement Number W911NF-13-2-0045 (ARL Cyber Security CRA). The views and conclusions contained in this document are those of the authors and should not be interpreted as representing the official policies, either expressed or implied, of the Combat Capabilities Development Command Army Research Laboratory or the U.S. Government. The U.S. Government is authorized to reproduce and distribute reprints for Government purposes not withstanding any copyright notation here on.

\clearpage
%
%

\end{document}